\documentclass[journal]{IEEEtran}
\ifCLASSINFOpdf
\else
\fi
\usepackage{algorithm}
\usepackage{algorithmic}

\usepackage{amsmath,amssymb}
\usepackage{graphicx}
\usepackage{caption}
\usepackage{subcaption}
\usepackage[keeplastbox]{flushend}
\usepackage{balance}

\usepackage{xcolor}
\newcommand{\new}[1]{\textcolor{black}{#1}}

\usepackage[all=normal,paragraphs=tight,floats=normal,mathspacing=normal,
wordspacing=tight,charwidths=tight,mathdisplays=normal,leading=normal]{savetrees}

\begin{document}
%
\title{Efficient channel charting via phase-insensitive \\ distance computation }
%
%
%

\author{Luc Le Magoarou
\thanks{Luc Le Magoarou is with Institute of Research and Technology b$<>$com, Rennes, France. Contact address: \texttt{luc.lemagoarou@b-com.com}.}}

%
%

\markboth{ACCEPTED VERSION}{}%

%



\maketitle

\begin{abstract}
Channel charting is an unsupervised learning task whose objective is to encode channels so that the obtained representation reflects the relative spatial locations of the corresponding users. It has many potential applications, ranging from user scheduling to proactive handover. In this paper, a channel charting method is proposed, based on a distance measure specifically designed to reduce the effect of small scale fading, which is an irrelevant phenomenon with respect to the channel charting task. A nonlinear dimensionality reduction technique aimed at preserving local distances (Isomap) is then applied to actually get the channel representation. The approach is empirically validated on realistic synthetic \new{multipath} MIMO channels, achieving better results than previously proposed approaches, at a lower cost.
\end{abstract}

\begin{IEEEkeywords}
channel charting, dimensionality reduction, MIMO signal processing, machine learning.
\end{IEEEkeywords}

%
\IEEEpeerreviewmaketitle

\section{Introduction}
\IEEEPARstart{M}{achine} learning techniques have been applied with success to several wireless communication problems in recent years \cite{Oshea2017}. However, most of these techniques fall within the supervised learning paradigm, and thus require labeled data to operate. The acquisition of such data may be unpractical or complex to implement within existing communication systems.

Channel charting \cite{Studer2018} on the other hand is a fully unsupervised learning task. Indeed, its objective is for a multi-antenna base station to build a low-dimensional map (called chart) of the radio environment based on uplink channel measurements, without requiring access of the users' actual locations. The chart should reflect as much as possible the physical reality, in the sense that the charting function should preserve spatial neighborhoods. Predicting this way the relative locations of users from channel measurements has many potential applications, ranging from \new{SNR prediction \cite{Kazemi2020} and pilot reuse \cite{Ribeiro2020}}, to  user grouping, proactive handover management or beam-finding (see \cite{Studer2018} for more details on potential applications).
\new{What makes channel charting particularly interesting compared to classical positioning methods is its fully unsupervised nature. Indeed, no link with the application layer in order to get locations from a global navigation satellite system (GNSS) is required (even offline to build a dataset). Only channel measurements are needed, which are readily accessible from the radio access network (RAN). Moreover, having access to the relative locations of users instead of their absolute locations is sufficient for most applications that need to assess the proximity of users. In that sense channel charting can be seen as an unsupervised alternative to radio maps \cite{Bi2019}.}

\noindent {\bf Contributions.} This paper proposes a computationally efficient method to perform channel charting, based on the computation of a distance measure that is designed to be insensitive to small scale fading and to locally reflect physical distance. This measure is theoretically motivated based on a simple physical channel model. It allows to compute a distance matrix from the training channels, which is then used within the Isomap \cite{Tenenbaum2000} nonlinear dimensionality reduction method to get the chart coordinates. The proposed method is empirically assessed on channel data used in \cite{Studer2018}, for which it yields better results than previously proposed approaches at a lower computational cost. The method is also assessed on higher-dimensional channels for which some of the previously proposed methods are too costly to be applied.

\noindent {\bf Related work.} In the seminal paper on channel charting \cite{Studer2018}, the raw second order moment of channels is used as input features in order to reduce the influence of small scale fading which is irrelevant to the channel charting task. Such features have the disadvantage of being of dimension equal to the square of the channel dimension. On the other hand, it was more recently proposed to use the channel autocorrelation as input features \cite{Gonultas2020}. This has the advantage of yielding features of the same dimension as the channel that are also quite insensitive to small scale fading. However, using autocorrelations automatically makes the features translation invariant in the angular and delay domains, which is a potentially harmful property with respect to the channel charting task, especially for channels in line of sight (LoS) or comprising a dominant path. In contrast, the method proposed in this paper does not require to square the channel dimension nor introduces any invariance in the angular or delay domain. Finally, it is interesting to notice that the distance measure used in this paper was also used for supervised learning tasks such as user positioning and channel mapping  in a previous work \cite{Lemagoarou2021}, although it was not theoretically motivated.


\section{Problem formulation}
The method proposed in this paper applies to a wide variety of multi-user massive multiple input multiple output (massive MIMO) wideband systems \cite{Rusek2013,Larsson2014}, operating indifferently in time division duplex (TDD) or frequency division duplex (FDD), where the antennas at the base station are indifferently colocated or not (in which case it is a distributed MIMO system). Let us consider $A$ base station antennas and $S$ subcarriers \new{evenly distributed at frequencies $f_1,\dots,f_S$ around a center frequency $f_c$ spanning a total bandwidth $B$}, and denote $\mathbf{h} \in \mathbb{C}^{AS}$ the uplink channel vector between any given user and the base station and $h_{a,s} \in \mathbb{C}$ the channel for the $a$th antenna on the $s$th subcarrier. In order to lighten notations, the total channel dimension is denoted $M\triangleq AS$. Note that no index is introduced to denote to which specific user corresponds the channel, since the proposed method treats indifferently the channels from all users.

Based on a database of $N$ estimated uplink channels 
\begin{equation}
\{\mathbf{h}_i\}_{i=1}^N,
\label{eq:dataset}
\end{equation} 
the objective in this paper is to build a forward charting function (or simply charting function)
\begin{equation}
\begin{array}{ccccl}
\mathcal{C} & : & \mathbb{C}^{M} & \to & \mathbb{R}^{D'} \\
 & & \mathbf{h} & \mapsto & \mathbf{z} \triangleq\mathcal{C}(\mathbf{h}), \\
\end{array}
\label{eq:charting_function}
\end{equation} 
where $\mathbf{z}$ is the vector in the chart associated to channel $\mathbf{h}$ and $D'$ is the chart dimension. Note that, as opposed to \cite{Studer2018}, no feature extraction step is considered, so that the forward charting function directly operates on estimated channels.

\noindent{\bf Performance measures.}
In order to evaluate charting functions, the location of the user yielding channel $\mathbf{h}_i$ is denoted $\mathbf{p}_i \in \mathbb{R}^D$, where $D$ is the number of considered spatial dimensions (two or three). A charting function $\mathcal{C}$ is considered good if it preserves neighborhoods. Mathematically, it corresponds to the property
\begin{equation}
\mathbf{p}_l \approx \mathbf{p}_k \Leftrightarrow \mathcal{C}(\mathbf{h}_l) \approx \mathcal{C}(\mathbf{h}_k)
\label{eq:good_property}
\end{equation}
being true for any $k,l$ using the Euclidean distance. In order to quantify this rather vague requirement, and following \cite{Studer2018}, the performance measures used in this paper are the continuity (CT) and trustworthiness (TW)\new{, which are classical performance measures for dimensionality reduction methods} \cite{Venna2001}. Both measures are between zero and one (the higher the better). Continuity assesses whether channels corresponding to nearby locations are mapped to nearby vectors in the chart (forward implication in \eqref{eq:good_property}). Its precise expression is
\begin{equation}
\text{CT}(K) = 1 - \tfrac{2}{NK(2N-3K-1)}\sum_{i=1}^N\sum_{j\in \mathcal{V}_i^K}\hat{r}(i,j)-K,
\label{eq:CT}
\end{equation}
where $\hat{r}(i,j)$ corresponds to the rank (in terms of proximity measured with the Euclidean distance) of the charted channel $\mathcal{C}(\mathbf{h}_j)$ with respect to $\mathcal{C}(\mathbf{h}_i)$ \new{($\hat{r}(i,j)=n$ if $\mathcal{C}(\mathbf{h}_j)$ is the $n$-th closest to $\mathcal{C}(\mathbf{h}_i)$ considering all the charted channels)}, and $\mathcal{V}_i^K$ is the set containing indices of channels being among the $K$ closest to the $i$th in space but not in the chart. \new{if the $K$ nearest neighbors of all considered channels in terms of spatial location are all mapped among the $K$ nearest neighbors of their representation on the chart, then $\text{CT}(K) = 1$. On the contrary, if the $K$ nearest neighbors of all considered channels in terms of spatial location are all mapped among the $K$ furthest of their representation on the chart, then $\text{CT}(K) = 0$.}  On the other hand, trustworthiness assesses whether nearby vectors in the chart do correspond to spatially close users (converse implication in \eqref{eq:good_property}). \new{It is defined in a very similar way as continuity, except that the roles of the spatial locations and locations on the chart are switched}. It is expressed as
\begin{equation}
\text{TW}(K) = 1 - \tfrac{2}{NK(2N-3K-1)}\sum_{i=1}^N\sum_{j\in \mathcal{U}_i^K}r(i,j)-K,
\label{eq:TW}
\end{equation}
where $r(i,j)$ corresponds to the rank (in terms of proximity measured with the Euclidean distance) of the location $\mathbf{p}_j$ with respect to  $\mathbf{p}_i$, and $\mathcal{U}_i^K$ is the set containing indices of channels being among the $K$ closest to the $i$th in the chart domain but not in space.

\section{Forward charting function}
In this section, the proposed forward charting function is introduced and explained in details. The general strategy is to first design a distance measure between channels that preserves spatial neighborhoods and then use it to compute chart coordinates via a nonlinear dimensionality reduction method.

\subsection{Distance measure}
In order to set up the proposed method, a distance measure $d$ between channels which locally reflects physical reality (locations of the corresponding users) is first sought.
If the Euclidean distance between channel vectors preserved spatial neighborhoods, no charting would be needed. However, this is not the case for several reasons. First of all, as noticed in \cite[Figure 4]{Studer2018}, using as distance measure
\begin{equation}
d(\mathbf{h}_k,\mathbf{h}_l)^2 =   \left\Vert\mathbf{h}_k-\mathbf{h}_l\right\Vert_2^2
\label{eq:dist_euclid}
\end{equation}
makes channels corresponding to users far away from the base station appear closer to one another and channels corresponding to users close to the base station appear further away to one another. This issue can be partly resolved by normalizing channels (for example simply dividing channel vectors by their $\ell_2$-norm \cite{Gonultas2020}), using the distance
\begin{equation}
d(\mathbf{h}_k,\mathbf{h}_l)^2 =   \left\Vert\frac{\mathbf{h}_k}{\left\Vert \mathbf{h}_k \right\Vert_2}-\frac{\mathbf{h}_l}{\left\Vert \mathbf{h}_l \right\Vert_2}\right\Vert_2^2.
\label{eq:dist}
\end{equation}

\begin{figure}[htbp]
\centering
\includegraphics[width=0.5\columnwidth]{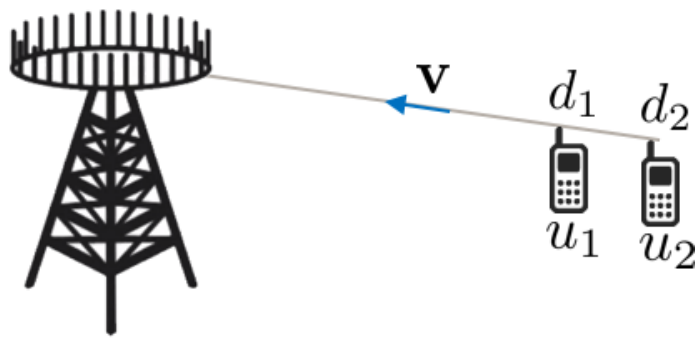}
\caption{Example of a pathological situation}
\label{fig:example}
\end{figure}

However, the Euclidean distance between normalized channels given in \eqref{eq:dist} is still unsatisfactory.
This can be seen with a very simple example exposing a pathological situation. Imagine two users $u_1$ and $u_2$ located in the same direction $\mathbf{v}$ with respect to the base station at distance $d_1$ and $d_2$ (as shown on figure~\ref{fig:example}).
Using the plane wave assumption, and considering a single propagation path, the normalized channels of users $u_1$ and $u_2$ can both be expressed as
\new{
\begin{equation}
\frac{\mathbf{h}_i}{\left\Vert \mathbf{h}_i \right\Vert_2} = \mathrm{e}^{-\mathrm{j}2\pi\frac{d_i}{\lambda}}\mathbf{f}(\tau_i)\otimes\mathbf{e}(\mathbf{v})
\end{equation}}
where \new{$\lambda \triangleq \frac{c}{f_c}$} is the wavelength \new{(at the central frequency)}, \new{$\tau_i \triangleq \frac{d_i}{c}$ is the time delay}, \new{$\mathbf{e}(\mathbf{v}) \in \mathbb{C}^A$} is the steering vector associated with direction $\mathbf{v}$ (see \cite{Sayeed2002,Lemagoarou2018} for their complete definitions) and \new{ $\mathbf{f}(\tau_i) \triangleq \frac{1}{\sqrt{S}}\left(\mathrm{e}^{-\mathrm{j}2\pi\tau_i(f_1-f_c)},\dots, \mathrm{e}^{-\mathrm{j}2\pi\tau_i(f_S-f_c)}\right)^T \in \mathbb{C}^S$ is the vector of relative phase differences along subcarriers associated to delay $\tau_i$ (simply equivalent to a steering vector in the frequency domain).} Now, if the two users are separated by half a wavelength, \new{assuming that the bandwidth is much smaller than the central frequency ($B\ll f_c$),} the two normalized channels are almost opposite vectors:
\new{\begin{equation}
|d_1-d_2| = \frac{\lambda}{2} \Rightarrow \left\Vert  \frac{\mathbf{h}_1}{\left\Vert \mathbf{h}_1 \right\Vert_2} - \frac{\mathbf{h}_2}{\left\Vert \mathbf{h}_2 \right\Vert_2} \right\Vert_2 \approx 2.
\end{equation}
This is because in that case, $\mathrm{e}^{-\mathrm{j}2\pi\frac{d_1}{\lambda}} = -\mathrm{e}^{-\mathrm{j}2\pi\frac{d_2}{\lambda}}$ and $\mathbf{f}(\tau_1)\approx \mathbf{f}(\tau_2)$.}
In other words, normalized channels are \new{almost} maximally far away while users are very close (a few centimeters at usual carrier frequencies). This simple example shows that the Euclidean distance between channels is very sensitive to small scale fading, which is clearly a harmful behavior with respect to the channel charting task. This sensitivity comes from the very fast changes in the global phase of channels due to small movements of users. In order to overcome this issue, a distance measure which is totally insensitive to the global phase of channels is proposed here:
\begin{equation}
d^\star(\mathbf{h}_k,\mathbf{h}_l)^2 \triangleq  \underset{\phi\in[0,2\pi]}{\text{min}} \left\Vert\frac{\mathbf{h}_k}{\left\Vert \mathbf{h}_k \right\Vert_2}-\mathrm{e}^{\mathrm{j}\phi}\frac{\mathbf{h}_l}{\left\Vert \mathbf{h}_l \right\Vert_2}\right\Vert_2^2.
\label{eq:dist_proposed}
\end{equation}
In the context of the previous example, this distance measure yields \new{$d^\star(\mathbf{h}_1,\mathbf{h}_2) = 0$ for $d_1=d_2$ and a monotonically increasing value of $d^\star$ with $|d_1-d_2|$ (for values of $|d_1-d_2|$ up to several meters for classical systems). This distance measure allows to get rid of the oscillating behavior (of period $\lambda$) of the distance given in \eqref{eq:dist} with respect to $|d_1-d_2|$ which is due to small scale fading}. 
The distance $d^\star$ may seem difficult to compute at first sight, because of the involved optimization problem. However, it can be expressed simply as
\begin{equation}
d^\star(\mathbf{h}_k,\mathbf{h}_l)^2 = 2 - 2\frac{| \mathbf{h}_k^H\mathbf{h}_l|}{\left\Vert \mathbf{h}_k \right\Vert_2\left\Vert \mathbf{h}_l \right\Vert_2}.
\label{eq:dist2}
\end{equation}
The proof of this equality is given in appendix~\ref{app:proof}. It means that taking the modulus of the inner product (instead of the real part) implicitly corresponds to put channels in phase (without needing to solve an optimization problem). Under this form, $d^\star$ is easy to compute. Note that the distance measure $d^\star$ is introduced here motivated by a simple example, but it is illustrated on realistic channels (comprising several paths) in the experimental section.

Compared to previously proposed channel charting methods, using this phase-insensitive distance $d^\star$ allows to gain insensitivity to small scale fading without introducing features of dimension $M^2$ (as proposed in \cite{Studer2018}), nor having to lose all sensitivity to absolute direction and delay (as is the case when using autocorrelations as input features \cite{Gonultas2020}).

\subsection{Chart coordinates}
Now that a distance measure has been chosen, the chart coordinates remain to be obtained. The objective is to find a global coordinate system in $D'$ dimensions (with $D'\ll M$) in which the Euclidean distances between training samples are as close as possible to the distances measured with $d^\star$ (that should be close to the spatial distance between users). To do so, a dimensionality reduction method can be applied.

It is interesting to notice that
distances measured with $d^\star$ reflect physical distances only locally (only small values of $d^\star$ are reliable). Indeed, a small value for $d^\star$ corresponds to nearby users, but a large value for $d^\star$ does not necessarily correspond to far away users. In such a situation, the Isomap method \cite{Tenenbaum2000} is particularly adapted. Indeed, it is based on the assumption that it is possible to reliably compute small distances between training samples but not large ones. It is based on a neighborhood graph (considering $k$ neighbors), from which large distances are estimated by finding shortest paths. In the studied context, this allows to approximate large distances using only small values of $d^\star$. Then, multidimensional scaling (MDS) \cite{Torgerson1952} is applied to the obtained distance matrix to get a low dimensional embedding (see \cite{Tenenbaum2000,Ghojogh2020} for a detailed description of Isomap). In the sequel, $\texttt{ISOMAP}(\mathbf{D},D',k)$ denotes the matrix whose columns are the result of the Isomap method applied to the input distance matrix $\mathbf{D}$ with an embedding in $D'$ dimensions considering the $k$ nearest neighbors to build the neighborhood graph. In practice, the scikit-learn \cite{scikit} implementation of Isomap is used. The proposed method is summarized in algorithm~\ref{alg:effcc}.

\begin{algorithm}[htb]
\caption{Proposed channel charting method}
\begin{algorithmic}[1] 
\REQUIRE Training channels $\{\mathbf{h}_i\}_{i=1}^N$, chart dimension $D'$
\STATE {\bf Distance computation:} \\ Build the matrix $ \mathbf{D} \in \mathbb{R}^{N\times N}$, with $ d_{ij} \leftarrow d^\star (\mathbf{h}_i,\mathbf{h}_j) $
\STATE {\bf Dimensionality reduction:} \\ $\mathbf{U}\in \mathbb{R}^{D' \times N} \leftarrow \texttt{ISOMAP}(\mathbf{D},D',k)$
\ENSURE $\mathcal{C}(\mathbf{h}_i) \leftarrow \mathbf{u}_i,\, i=1,\dots, N$ (chart coordinates)
\end{algorithmic}
\label{alg:effcc}
\end{algorithm}

\noindent {\bf Computational complexity.} One of the main advantages of the proposed method compared to prior art is its low complexity with respect to the channel dimension $M$ and number of training channels $N$. Indeed, the distance computation step has complexity $\mathcal{O}(MN^2)$, while the dimensionality reduction step has complexity $\mathcal{O}(N^2\log N)$ \cite{Silva2002,scikit}. Note that the original channel charting methods based on raw second order moments \cite{Studer2018} have complexity at least $\mathcal{O}(M^2N^2)$, or require to train a neural network (whose complexity is difficult to precisely quantify but which takes time in practice).

In summary, the proposed method is particularly adapted to high-dimensional channels (large $M$), with relatively few examples (not too large $N$). However, if a lot of examples are available, one can perfectly envision to use landmark Isomap \cite{Bengio2003} instead of the classical Isomap. This would allow to reduce complexity of the method to $\mathcal{O}(Mn^2)$  for the first step and $\mathcal{O}(nN\log N)$ for the second, with $n\ll N$ being the number of considered landmarks. 

\section{Experiments}
In this section, the proposed method is empirically assessed on several kinds of MIMO channels. At first, and in order to compare to previously proposed methods \cite{Studer2018,Huang2019}, training channels obtained with the Quadriga channel simulator \cite{Jaeckel2014} are used. Then, higher dimensional channels taken from the DeepMIMO dataset \cite{Alkhateeb2019} are considered. \new{In all the following experiments, the dimension of the chart is fixed to $D'=2$. Besides the obvious advantage of this choice for visualizing the chart, it comes from the fact that if users are distributed on a 2D-plane, then the channel manifold should be two-dimensional (see \cite[Section III.C]{Lemagoarou2021} for a more detailed explanation of this point). One could of course optimize the chart dimension empirically. This option is left to future work.}

\subsection{Quadriga channels}
\begin{figure}[htbp]
\centering
\includegraphics[width=0.7\columnwidth]{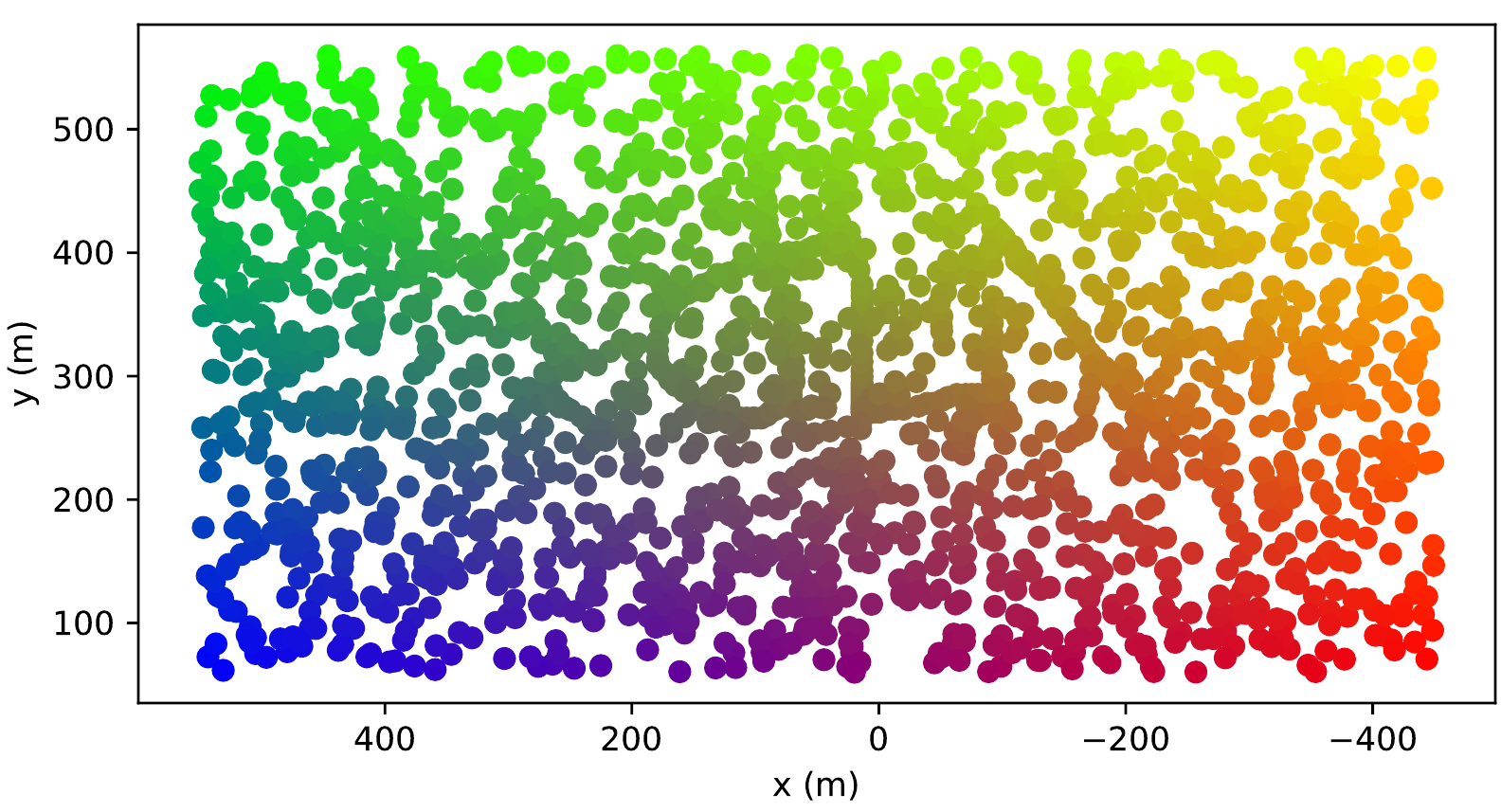}
\caption{User locations for Quadriga channels}
\label{fig:users_quadriga}
\end{figure}

For this first set of experiments, exactly the same \new{multipath} channels as those used in \cite{Studer2018} are considered. Namely, $N = 2048$ users are randomly located in an area of $1000\,\text{m} \times 500\,\text{m}$ and their channels are obtained considering an urban macrocell environment at a center frequency of $2\,\text{GHz}$ within the Quadriga channel simulator \cite{Jaeckel2014} (details available in \cite[Table I]{Studer2018}). User locations are shown on figure~\ref{fig:users_quadriga}, the base station being at the origin and equipped with an uniform linear array (ULA) of $32$ half-wavelength separated antennas. The dimension of these training channels is $M=32$, and they are assumed to be acquired at an SNR of $0\,\text{dB}$ and averaged over $10$ time instants.

\begin{figure}[htbp]
\centering
\includegraphics[width=0.5\columnwidth]{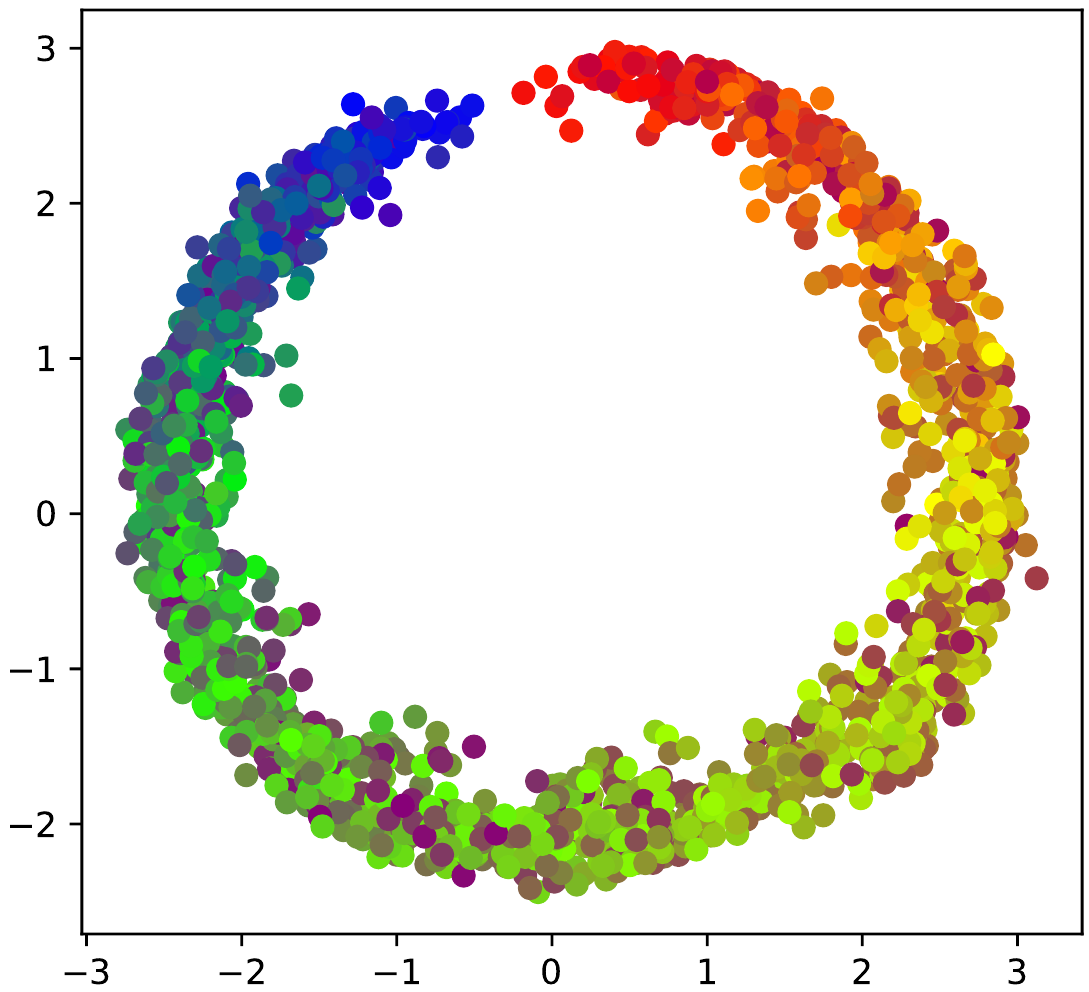}
\caption{Obtained chart for Quadriga channels (nLoS)}
\label{fig:chart_quadriga}
\end{figure}

\noindent {\bf Results.} The obtained chart for $D'=2$ and $k=30$ is shown on figure~\ref{fig:chart_quadriga} (in the nLoS case). Neighborhoods are quite well preserved, although channels seem to be mapped on a circle. This may be explained by the fact that using a single subcarrier makes the system insensitive to delay, so that only angle discrimination is possible (no distance information appears on the chart). Results in terms of continuity (CT) and trustworthiness (TW) are shown on the leftmost (LoS) and center (nLoS) parts of figure~\ref{fig:res}, where \new{it} is compared to the original channel charting methods of \cite{Studer2018} (PCA, SM, SM+ and AE, see \cite{Studer2018} for the details of each method). The obtained performance is very good. Indeed, the proposed method outperforms previously proposed methods in terms of continuity and is at the same level in terms of trustworthiness, while being more computationally efficient. Indeed, it does not require to compute high dimensional features, which is the case for the concurrent methods which use raw second order moments of dimension $M^2$. The proposed method runs in less than $6$ seconds on a regular laptop (Intel(R) Core(TM)
i5-6300U CPU @ 2.40 GHz), which makes it very fast. Moreover, note that the proposed method is also better than the previously proposed method that uses channel autocorrelation as input features (see \cite[Table II]{Huang2019}, ``Plain'' column to precisely compare).

\subsection{DeepMIMO channels}
\begin{figure}[htbp]
\centering
\includegraphics[width=0.6\columnwidth]{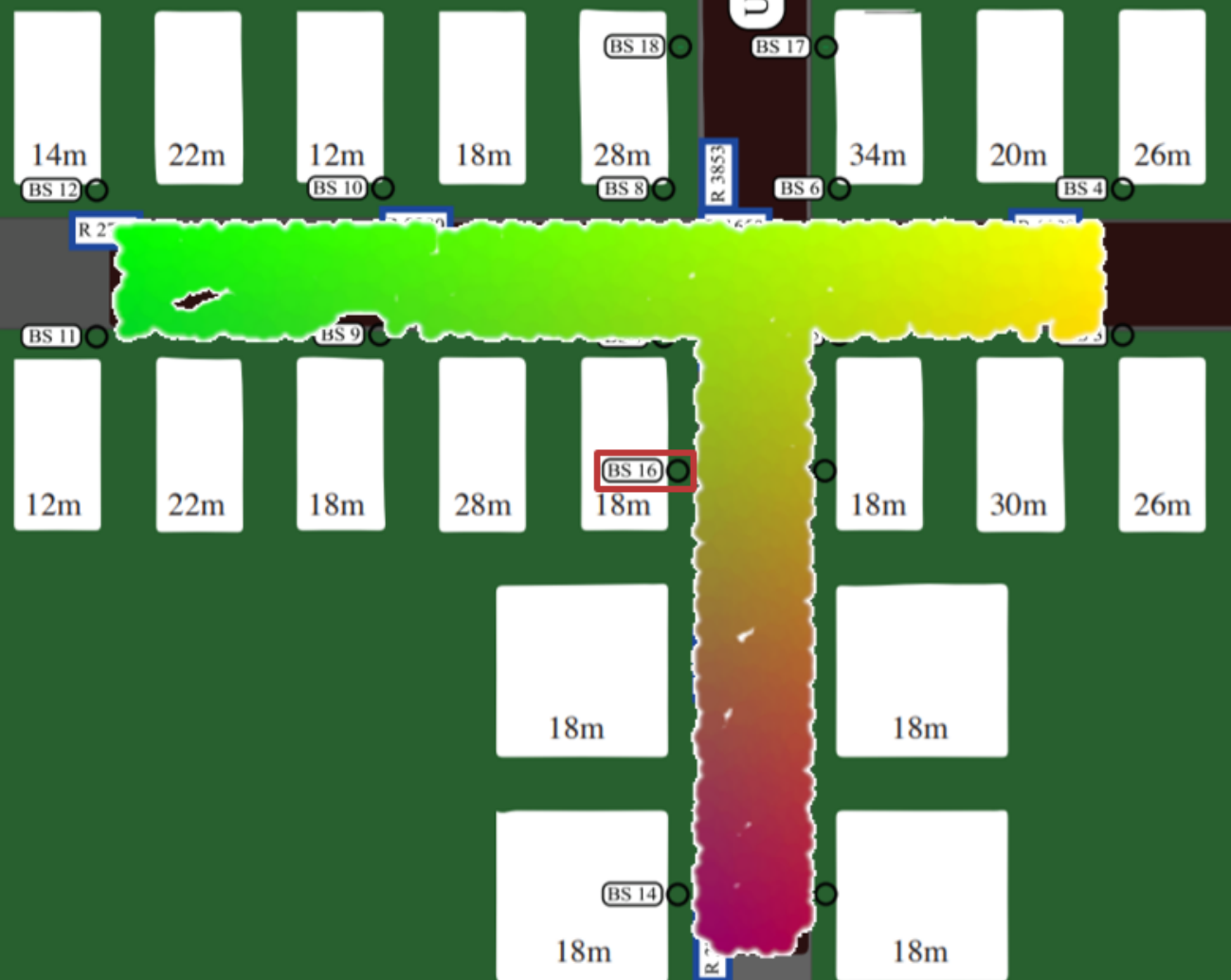}
\caption{User locations for DeepMIMO channels}
\label{fig:users_deepmimo}
\end{figure}

For this second experiment, $N=3000$ training \new{multipath} channels are taken from the DeepMIMO dataset \cite{Alkhateeb2019}. The `O1' urban outdoor ray-tracing scenario is chosen (with $5$ paths per channel), with a single base station (BS $16$) and a subset of all possible user locations. The scenario is depicted on figure~\ref{fig:users_deepmimo}, where the colorized points correspond to users whose channels are in the training data, and the base station is circled in red. The base station is equipped with a square uniform planar array (UPA) with $64$ half-wavelength separated antennas at a central frequency of $3.5\,\text{GHz}$, with $16$ subcarriers evenly spaced spanning a band of $20\,\text{MHz}$. This results in channels of dimension $M=1024$. No noise is added for this experiment. \new{Note that, as opposed to the previous experiment, only a single time instant is considered here for the channels (no averaging is done).}

\begin{figure}[htbp]
\centering
\includegraphics[width=0.5\columnwidth]{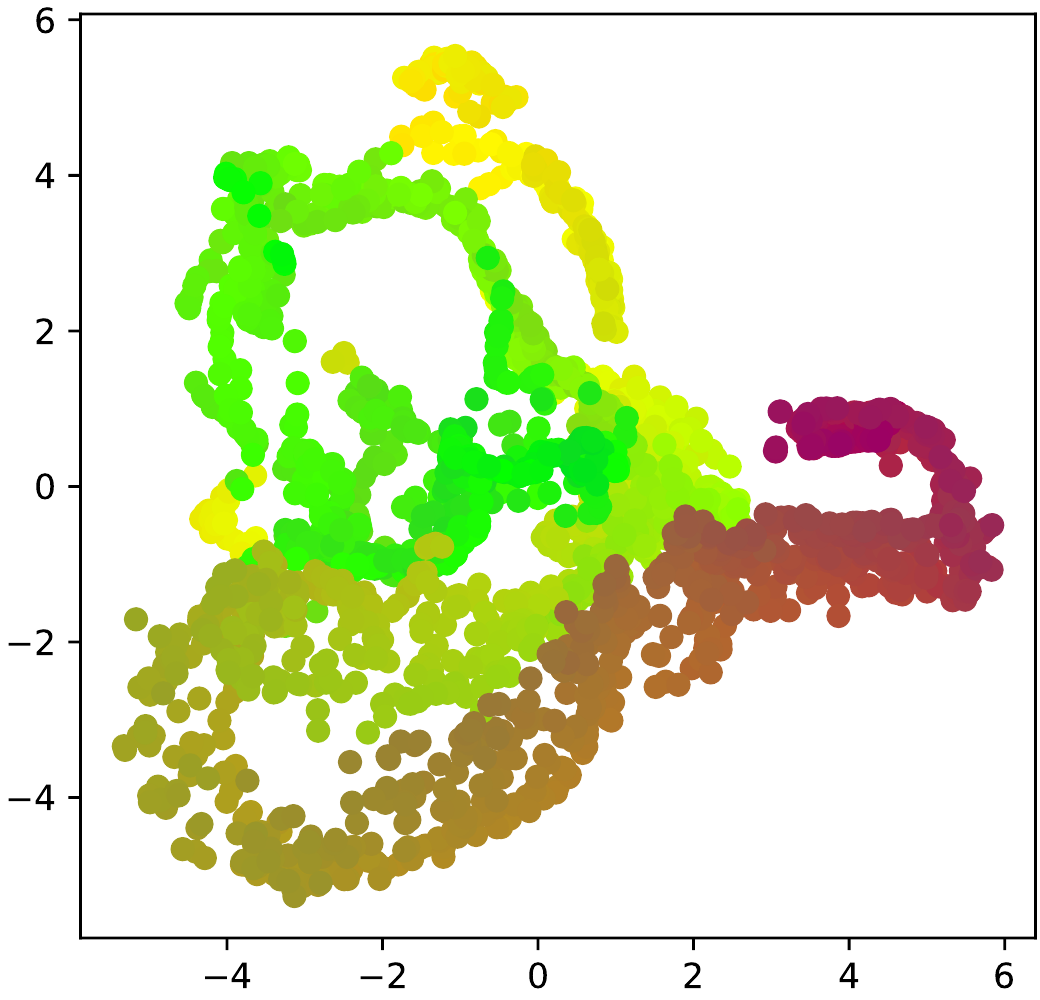}
\caption{Obtained chart for DeepMIMO channels}
\label{fig:chart_deepmimo}
\end{figure}

\begin{figure*}[t!]
\centering
\includegraphics[width=0.329\textwidth]{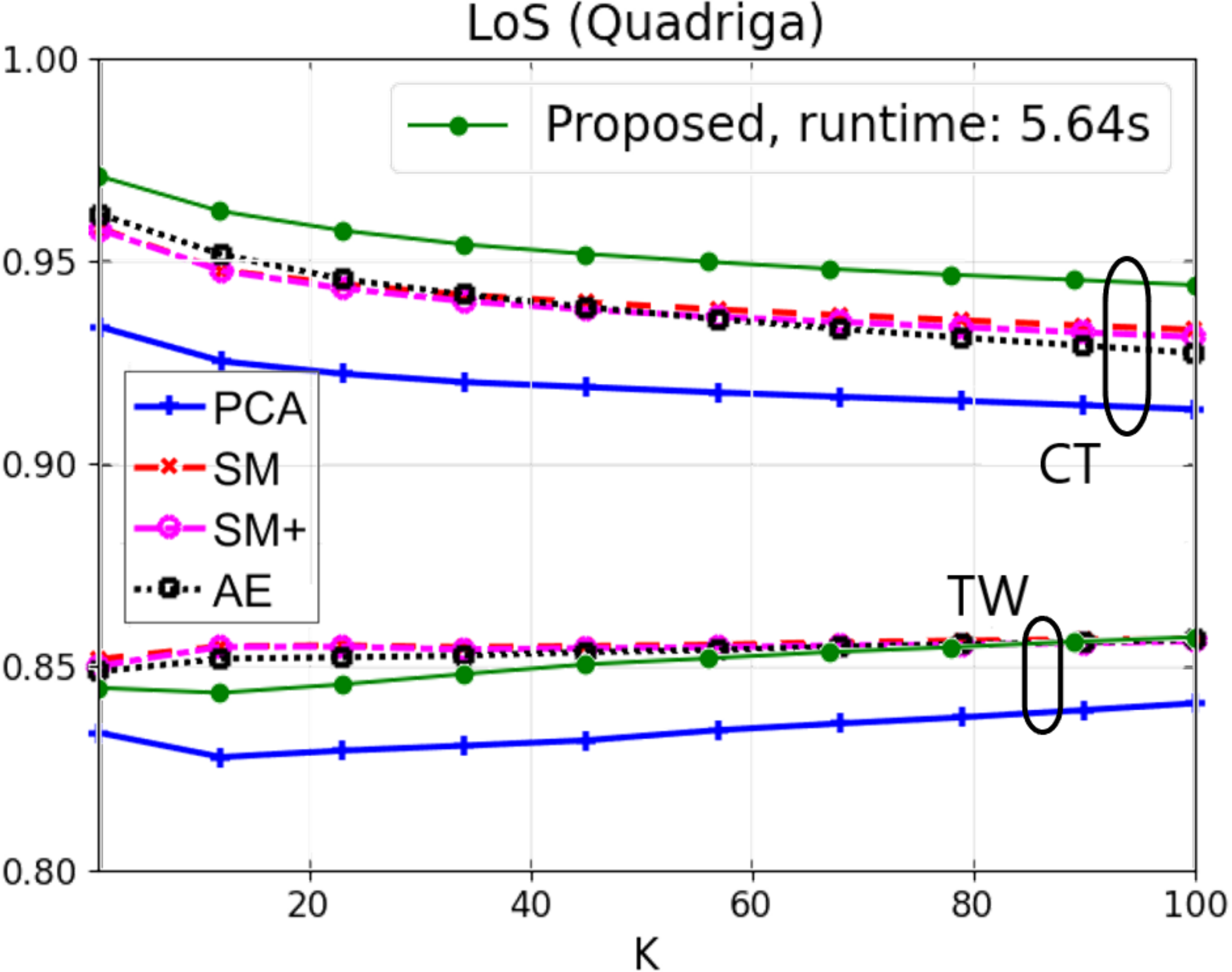}
\includegraphics[width=0.329\textwidth]{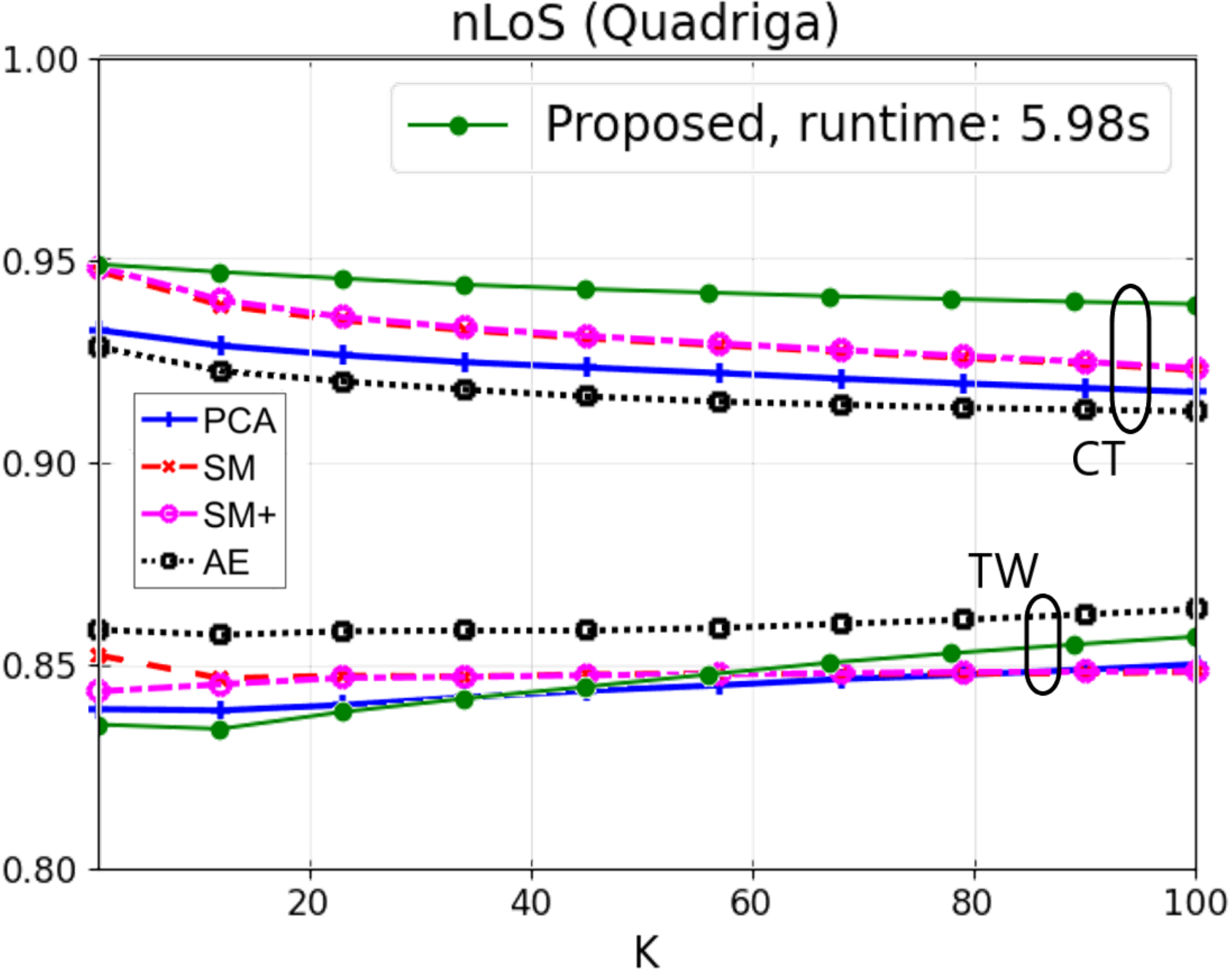}
\includegraphics[width=0.329\textwidth]{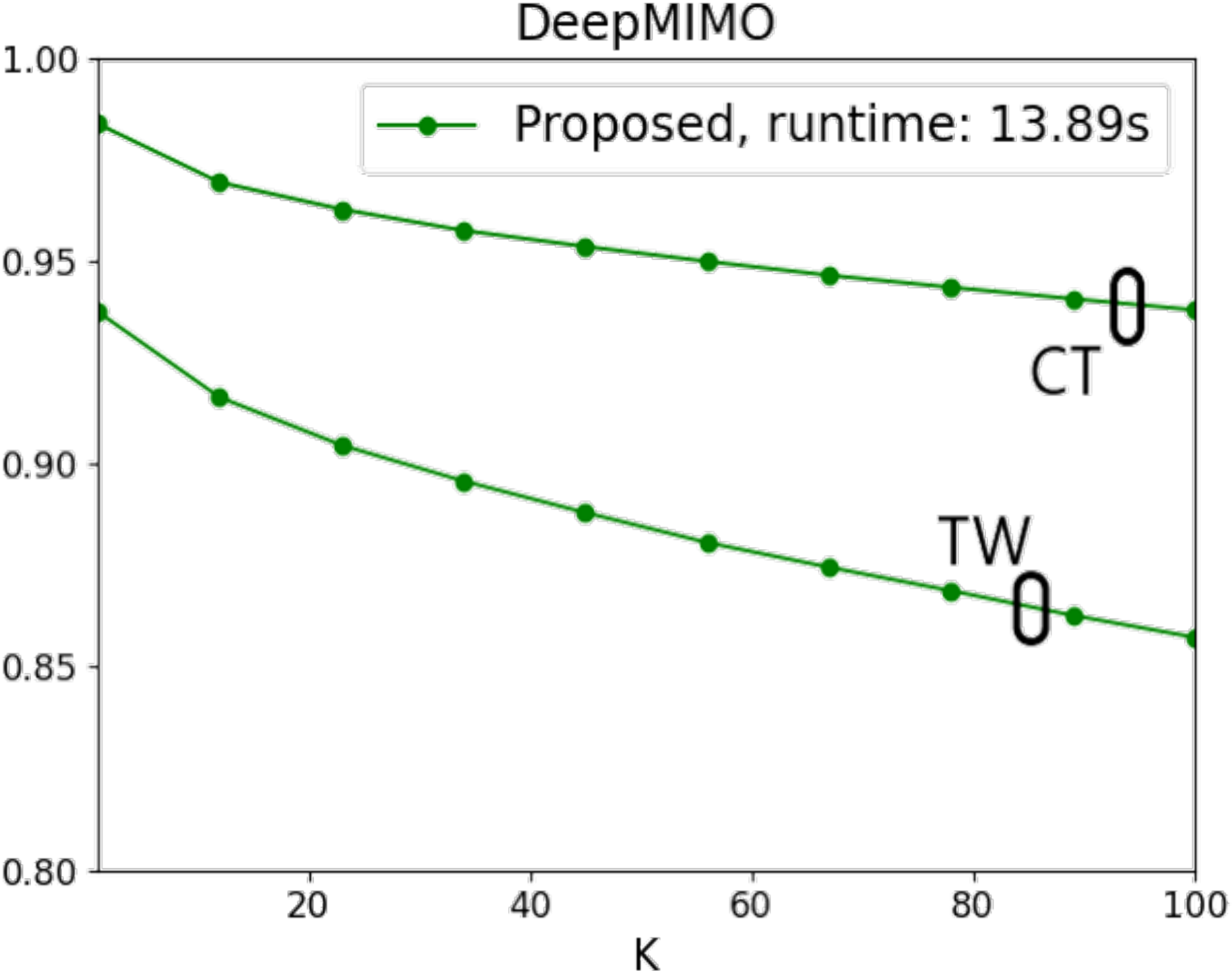}
\caption{Continuity (CT) and trustworthiness (TW) results for the Quadriga and DeepMIMO channels. The \new{proposed} method is compared to several baselines proposed in \cite{Studer2018}.}
\label{fig:res}
\end{figure*}

\noindent {\bf Results.} The obtained chart for $D'=2$ and $k=30$ is shown on figure~\ref{fig:chart_deepmimo}, where it is seen that spatial neighborhoods are well preserved. In particular, thanks to the fact that several subcarriers are considered \new{(as opposed to the previous experiment)}, the system exhibits sensitivity to delay, so that both the angle and distance information are expressed in the chart. Regarding the CT and TW performance measures for a varying number of neighbors $K$, results are shown on the rightmost part of figure~\ref{fig:res}. The performance seems better than for the Quadriga channels, which may be explained by the fact that channels are noiseless. More interestingly, the proposed charting method runs in less than $15$ seconds on a regular laptop, even though channels are high-dimensional ($M=1024$). This is promising for its applicability, since concurrent methods \cite{Studer2018,Huang2019} would take much more time for such channels. For example, the raw second order moment for such channels would comprise $M^2 = 2^{20}$ entries, this is why methods of \cite{Studer2018} are not compared to the proposed one for these channels.

\section{Conclusion and perspectives}
In this paper, a computationally efficient channel charting method was introduced. It is based on the computation of a distance measure between channel vectors which is designed to be insensitive to small scale fading. The distance matrix obtained with this measure is then used within the Isomap nonlinear dimensionality reduction method in order to obtain the chart coordinates. The proposed method is assessed on realistic \new{multipath} channels of various dimensions, showing great potential in terms of accuracy and computational efficiency.

In the future, several research leads could be investigated in the continuation of this paper. First of all, it would be interesting to handle out of sample channels in order to allow for online implementations in practical systems. 
Second, the method could be used as an initialization for a deep neural network in order to perform fine tuning (as done for example in \cite{Lemagoarou2021} for channel mapping and user positioning). 


\appendices

\section{Proof of equation \eqref{eq:dist2}}
\label{app:proof}
The proof starts with a re-expression of the norm as
$$
 \left\Vert\frac{\mathbf{h}_i}{\left\Vert \mathbf{h}_i \right\Vert_2}-\mathrm{e}^{\mathrm{j}\phi}\frac{\mathbf{h}_j}{\left\Vert \mathbf{h}_j \right\Vert_2}\right\Vert_2^2 =  2 - 2\frac{\mathfrak{Re}\left\{ \mathbf{h}_i^H\mathrm{e}^{\mathrm{j}\phi}\mathbf{h}_j\right\}}{\left\Vert \mathbf{h}_i \right\Vert_2\left\Vert \mathbf{h}_j \right\Vert_2}.
$$
\new{Then, using $\mathfrak{Re}(z)\leq |z|$ and noticing that
$
\mathfrak{Re}\left\{ \mathbf{h}_k^H\mathrm{e}^{\mathrm{j}\phi}\mathbf{h}_l\right\} = \left| \mathbf{h}_k^H\mathbf{h}_l \right| $ for $\phi = \arctan\left[\frac{\mathfrak{Re}\left\{ \mathbf{h}_k^H(\mathrm{j}\mathbf{h}_l)\right\}}{\mathfrak{Re}\left\{ \mathbf{h}_k^H\mathbf{h}_l\right\}}\right]$ concludes the proof.}
\ifCLASSOPTIONcaptionsoff
  \newpage
\fi



\bibliographystyle{IEEEtran}
\bibliography{biblio}
\end{document}